\documentclass[journal,twoside,web]{ieeecolor}
\usepackage[table,xcdraw]{xcolor}

\usepackage{tmi}
\usepackage{cite}
\usepackage{amsmath,amssymb,amsfonts}
\usepackage{algorithmic}
\usepackage{graphicx}
\usepackage{textcomp}
\usepackage[hidelinks]{hyperref}
\usepackage{cleveref}
\usepackage{multirow}
\usepackage{booktabs}
\usepackage{subcaption} 
\usepackage{dsfont}

\usepackage{colortbl}
\usepackage{stfloats}

\usepackage{fontawesome5}
\usepackage{diagbox}
\usepackage{rotating}
\usepackage{threeparttable}

\def\ie{\emph{i.e.}}
\def\eg{\emph{e.g.}}

\def\etal{\emph{et al.}}

\newcommand{\figref}[1]{Figure~\ref{#1}}
\newcommand{\tabref}[1]{Table~\ref{#1}}

\newcommand{\secref}[1]{\S\ref{#1}}

\definecolor{mygray4}{gray}{0.8}
\definecolor{mygray5}{gray}{0.93}
\definecolor{myGreen2}{RGB}{236, 246, 239}

\definecolor{myGreen}{RGB}{15, 157, 88}
\newcommand{\tmyC}[1]{\textcolor{blue}{#1}}
\newcommand{\tmyB}[1]{\textcolor{myGreen}{#1}}
\newcommand{\tmyA}[1]{\textcolor{red}{#1}}

\def\ourproject{\textsc{Colon-X}}
\def\ourdata{\textsc{ColonVQA}}
\def\ourdataE{\textsc{ColonEval}}
\def\ourdataP{\textsc{ColonPert}}
\def\ourdataR{\textsc{ColonReason}}
\def\ourmodel{\textsc{ColonR1}}

\AtBeginDocument{%
  \definecolor{lightblue}{rgb}{0.678,0.847,0.902}%
}

\usepackage{pifont}

\usepackage{siunitx}
\usepackage{makecell}
\def\BibTeX{{\rm B\kern-.05em{\sc i\kern-.025em b}\kern-.08em
    T\kern-.1667em\lower.7ex\hbox{E}\kern-.125emX}}
\begin{document}

\title{\ourproject: Advancing Intelligent Colonoscopy toward Clinical Reasoning}
\author{Ge-Peng Ji, Jingyi Liu, Deng-Ping Fan, Huazhu Fu, and Nick Barnes
\thanks{Corresponding author: Deng-Ping Fan (dengpfan@gmail.com) is with NKIARI \& SLAI, Shenzhen Futian and VCIP, Nankai University, Tianjin, China. Ge-Peng Ji and Nick Barnes are with the Australian National University (ANU), Canberra, Australia. Jingyi Liu is with the King Abdullah University of Science and Technology (KAUST), Thuwal, Saudi Arabia. Huazhu Fu is with the Institute of High Performance Computing
(IHPC), Agency for Science, Technology and Research (A*STAR), Singapore.}
\thanks{This work was supported in part by the National Natural Science Foundation of China (No.62476143), and the Fundamental Research Funds for the Central Universities (Nankai University, No.63243150). We sincerely thank certified endoscopists from the Third People's Hospital of Jingzhou for their valuable time and expertise in reviewing our data.}
}

\IEEEaftertitletext{
\centering
\vspace{-2em} 
\includegraphics[width=\linewidth]{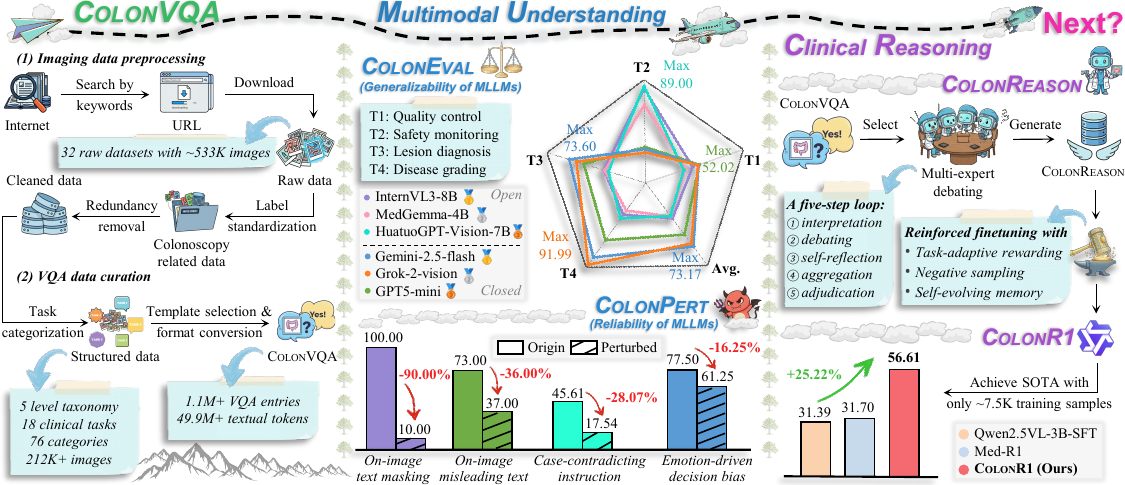}
\captionof{figure}{\textbf{Research roadmap of {\ourproject} project.} 
Building upon the most comprehensive multimodal colonoscopy database ({\ourdata} as in \secref{sec:colonvqa}), we propel a pivotal transition in intelligent colonoscopy, evolving from \textit{multimodal understanding} ({\ourdataE} in \secref{sec:coloneval} \& {\ourdataP} in \secref{sec:colonpert}) to \textit{clinical reasoning} ({\ourdataR} in \secref{sec:colonreason} \& {\ourmodel} in \secref{sec:colonr1}). These efforts collectively illuminate the path to next-generation advances in colonoscopy and broader medical applications.}
\label{fig:teaser_figure}
\vspace{2em} 
}

\maketitle


\maketitle

\begin{abstract}
In this study, we present {\ourproject}, an open initiative aimed at advancing intelligent colonoscopy toward clinical reasoning.
We begin by constructing {\ourdata}, the most comprehensive multimodal dataset ever built for colonoscopy, featuring over 1.1 million visual question answering entries across 76 clinical findings and 18 multimodal tasks.
Beyond serving as a community-wide data foundation, we investigate a critical yet underexplored transition:
(a) To capture the current landscape of multimodal understanding behaviors, we systematically assess the generalizability of 22 multimodal large language models in colonoscopy and examine their reliability under human-induced perturbations. The results reveal that clinical outputs from leading MLLMs remain far from robust and trustworthy.
(b) To narrow this gap, we further explore reasoning-centric intelligence tailored for colonoscopy. Specifically, we curate {\ourdataR}, a clinically grounded reasoning dataset
annotated through a multi-agent debating pipeline, and develop {\ourmodel}, the first R1-styled model that mitigates reward information collapse through task-adaptive rewards and gradient-stable policy optimization.
Under data-scarce conditions, our {\ourmodel} achieves 56.61\% overall accuracy, outperforming supervised fine-tuning by 25.22\%, and sets a new reasoning-enabled baseline for multimodal colonoscopy analysis.
All data and model resources are publicly available at \url{https://github.com/ai4colonoscopy/Colon-X}.
\end{abstract}

\begin{IEEEkeywords}
Colonoscopy, Multimodal Understanding, Multimodal Reasoning, and Abdomen.
\end{IEEEkeywords}

\section{Introduction}

Colonoscopy, the gold standard for early colorectal cancer detection \cite{ENG2024}, remains limited by operator variability and fatigue, underscoring the need for intelligent colonoscopy \cite{ji2026frontiers}. Recent studies indicate that (semi-)automated assistance reduces the miss rate of colorectal neoplasia by nearly 50\%, compared to conventional workflows \cite{wallace2022impact}. Yet, intelligence in colonoscopy still trails behind its expectations in the general domain, especially in multimodal topics \cite{guo2025deepseekr1,achiam2023gpt}. This raises a crucial question: 
\textit{``Have multimodal models in colonoscopy truly progressed beyond understanding to clinically grounded reasoning?''}
To embrace this challenge, we launch the \textbf{\ourproject} project, an open initiative aimed at advancing multimodal intelligence in colonoscopy and beyond.

As shown in \figref{fig:teaser_figure}, we begin by introducing {\ourdata}, the most extensive database ever built for multimodal colonoscopy analysis, featuring 1,100,786 visual question answering (VQA) queries, equivalent to over 49.9 million textual tokens.
It is distinguished by its \textit{category-rich} composition, containing 212,742 images across 76 clinically meaningful findings, and \textit{task-diverse} design, covering 18 multimodal tasks organized within a five-level taxonomy. Together, this foundation drives community-wide progress to a pivotal transition from understanding to reasoning. 

As the first step in such a transition, we characterize current model behaviors in \textit{multimodal understanding} along two essential but understudied dimensions.

\noindent$\bullet$ \textbf{Generalizability} (\secref{sec:coloneval}) {\faHandPointRight[regular]} We introduce a clinically reviewed set, \underline{\ourdataE}, that assesses the generalizability of 22 multimodal large language models (MLLMs) across diverse colonoscopy tasks. Our evaluation yields three key observations. (a) Performance gap: closed-source MLLMs hold overall superiority, but open-source models exhibit advantages in sub-tasks like safety monitoring. (b) Specialist's paradox: for open-source models, some generalists unexpectedly surpass medical-specific variants, questioning the current training strategies for task specialization. (c) Reasoning-outcome gap: for closed-source models, reasoning-enabled variants tend to enhance clinical interpretability but not necessarily decision-making accuracy. 

\noindent$\bullet$ \textbf{Reliability} (\secref{sec:colonpert}) {\faHandPointRight[regular]} Further, we introduce \underline{\ourdataP}, a test suite to quantify the robustness of leading MLLMs against human-induced perturbations. Given the uniqueness of colonoscopy data, we identify two forms of text-dominance bias that compromise clinical reliability: (a) implicit bias, triggered by manipulating on-image text, \ie, masking text embedded in images or replacing it with misleading text; and (b) explicit bias, caused by case-contradicting descriptions or emotionally-charged expressions within textual instructions.

Although large reasoning models (\eg, o-series \cite{jaech2024openaio1,o4mini}, DeepSeek-R1 \cite{guo2025deepseekr1}) have demonstrated impressive chain-of-thought capability in complex tasks \cite{guo2024deepseekcoder,yang2025qwen3}, their potential in colonoscopy remains largely unexplored. This drives us to advance this frontier toward \textit{\textbf{clinical reasoning}} through both data and model innovations.

\noindent$\bullet$ \underline{\textbf{\ourdataR}} (\secref{sec:colonreason}) {\faHandPointRight[regular]}
Obtaining human-annotated reasoning traces is costly and subjective across observers, making scalable reasoning annotation impractical. Thereby, we design a multi-agent debating pipeline that generates clinically grounded reasoning traces.
Human verification suggests that these synthesized traces are reasonably consistent with human judgment, supporting their use as structured supervisory signals for building a reasoning model.

\noindent$\bullet$ \underline{\textbf{\ourmodel}} (\secref{sec:colonr1}) {\faHandPointRight[regular]}
We observe severe optimization instability when applying naive GRPO \cite{guo2024deepseekcoder} to colonoscopy tasks, even with careful hyperparameter tuning. We attribute this to \textit{reward information collapse}, where the reward fails to preserve discriminative signals across heterogeneous task structures and case distributions.
To address this, we propose {\ourmodel}, the first R1-style framework reinforced on colonoscopy reasoning data featuring three key innovations.
First, unlike generic binary rewards, we introduce a task-adaptive reward scheme that maintains discrimination across diverse task structures.
Second, the imbalanced distribution of case difficulty often leads to advantage vanishing, as intra-group samples receive identical rewards under all-correct (easy) or all-incorrect (hard) conditions.
We mitigate this issue in two complementary ways: (a) negative sampling to restore contrast in easy cases, and (b) a self‑evolving memory mechanism that leverage past errors to guide future rollouts, preventing persistent failures on hard cases when their advantage signals vanish.
Trained on only $\sim$7.5K samples, our reinforced model outperforms the supervised fine-tuning counterparts by over 20\% on {\ourdataE} (see \tabref{tab:colon_r1}).

In summary, we fill a longstanding gap in colonoscopy by bringing together dataset ({\ourdata} \& {\ourdataR}), evaluation ({\ourdataE} \& {\ourdataP}), and methodology ({\ourdataR}) within a unified multimodal framework. 

\section{Related Works}

Over the past decade, intelligent colonoscopy \cite{fan2020pranet,ji2022video,ji2026frontiers} has progressed rapidly, driven by the emergence of various dedicated benchmarks. They can be broadly categorized into three groups based on distinct task objectives.

\noindent\textbf{Low-level vision tasks} are crucial in supporting reliable downstream analysis, where benchmark development has advanced along two directions. The first focuses on signal restoration, including super-resolution \cite{almalioglu2020endol2h,chen2022dynamic}, denoising \cite{zou2019cnn}, deblurring \cite{ali2021deep,queiroz2019endoscopy}, illumination correction \cite{bai2024endouic}, and specular reflection removal \cite{sanchez2017bright}. The second centers on extracting basic features, with benchmarks targeting edge detection \cite{tajbakhsh2015automated}, texture/color enhancement \cite{mathew2022clts}, depth estimation \cite{bobrow2023colonoscopy}, and perceptual quality assessment \cite{yue2023perceptual}.

\noindent\textbf{High-level vision tasks} focus on semantic interpretation of clinical findings in colonoscopy.
Fan {\etal} \cite{fan2020pranet} introduced a seminal benchmark that catalyzed polyp segmentation research. Since then, benchmark development has followed two directions.
On one direction, breadth oriented efforts have broadened the scope of clinical tasks. The Kvasir series exemplifies this trend, extending to multi-class gastrointestinal disease detection \cite{pogorelov2017kvasir,Borgli2020,smedsrud2021kvasir} and instrument segmentation \cite{jha2021kvasir}. Other benchmarks targeted bowel preparation \cite{pogorelov2017nerthus} and safety monitoring \cite{mesejo2016computer,jha2024polypdb}, reflecting growing attention to procedural risk assessment.
Other direction emphasizes depth by offering finer granularity of clinical findings. SUN-database \cite{misawa2021development} enriched lesion labels with attributes (\eg, polyp size, morphology) to support explainable diagnosis. Building on this, SUN-SEG \cite{ji2022video} introduced dense temporal masks, building the first large-scale benchmark for video polyp segmentation.

\begin{table}[t!]
\centering
\scriptsize
\renewcommand{\arraystretch}{1.2}
\renewcommand{\tabcolsep}{1.35mm}
\caption{\textbf{Overview of existing multimodal benchmarks related to colonoscopy.} We provide the count of classes (CLS), tasks (TSK), and VQA entries. The last four columns indicate support for multi-center sources (MS), multi-granularity labels (ML), perturbation testing (PT), and reasoning (RE).}
\label{tab:comp_of_multimodal_benchmarks}
\begin{tabular}{r || c | rrr | cccc}
\hline
Benchmark name & Year  &CLS &TSK & VQA &MS &ML &PT &RE \\
\hline
Kvasir-VQA \cite{gautam2024kvasirvqa} & 2024 & 5 & 6 & 58,849
& & & & \\
Kvasir-VQA-x1 \cite{gautam2025kvasirvqax1} & 2025 & 5 & 18 & 159,549
& & \checkmark &\checkmark & \\
EndoVQA-Instruct \cite{liu2025endobench} &2025 &4   &12 &439,703 &\checkmark &\checkmark &  & \\
EndoBench \cite{liu2025endobench} & 2025 & 4 & 12 & 6,832
& \checkmark & \checkmark & & \\
Gut-VLM \cite{khanal2025gutvlm} & 2025 & 8 & 12 & 21,792
& & &\checkmark & \\
ColonINST \cite{ji2026frontiers}
& 2025 & 62 & 4 & 450,724
& \checkmark & \checkmark & & \\
\hline
\rowcolor{myGreen2}
\textbf{\ourproject~(Ours)} 
& - & \textbf{76} & \textbf{18} & \textbf{1,100,786}
& \checkmark & \checkmark & \checkmark & \checkmark \\
\hline
\end{tabular}
\end{table}

\noindent\textbf{Multimodal tasks} are an emerging frontier focused on interpreting multimodal inputs (specifically vision-language modalities in this study) during colonoscopy. 
\tabref{tab:comp_of_multimodal_benchmarks} summarizes several recent multimodal benchmark related to colonoscopy. Kvasir-VQA \cite{gautam2024kvasirvqa} constructed 58.8K+ VQA entries based on 6.5K images in \cite{Borgli2020,jha2021kvasir}. Kvasir-VQA-x1 \cite{gautam2025kvasirvqax1} further expands this to 159.5K+ pairs, designed to assess MLLMs under imaging degradation via simple visual augmentations like brightness adjustments. Liu \etal~\cite{liu2025endobench} integrate 21 gastrointestinal datasets into 439K+ VQA entries, and further curate EndoBench, a well-designed 6,832-sample subset for MLLM evaluation. Gut-VLM \cite{khanal2025gutvlm} leveraged GPT-4o to generate 21.8K+ VQA pairs from 1,816 images \cite{pogorelov2017kvasir}, with a focus on exploring hallucination issues in MLLMs. Notably, above benchmarks primarily focus on endoscopic scenes, including findings beyond human colon such as esophagus and stomach. ColonINST \cite{ji2026frontiers} is the first multimodal dataset dedicated for colonoscopy, which comprises over 303K+ images and 450.7K+ VQA entries across four clinical tasks, 62 categories for domain-specific instruction-tuning.

\noindent\textbf{Remarks.} We mainly focus on clinical findings captured from the lower gastrointestinal tract, ensuring comprehensive coverage of colonoscopy scenarios. Compared to concurrent benchmarks as in \tabref{tab:comp_of_multimodal_benchmarks}, {\ourproject} offers the most extensive, category-rich, and task-diverse database ever built, establishing a solid data foundation to drive the next wave of in colonoscopy. Beyond this million-scale dataset, we delve into an essential but underexplored transition that evolves from multimodal understanding (generalizability and reliability) toward clinical reasoning (data and model) in colonoscopy.

\section{Scaling Colonoscopy Data to Million Scale}\label{sec:colonvqa}

Currently, the field still struggles with a persistent benchmarking crisis \cite{mahmood2025benchmarking}, which stems not only from the scarcity of biomedical data, but also from the convention of task-specific models trained on isolated benchmarks.
To address this, we construct {\ourdata} by consolidating public data sources, thus
enabling task-modality synergies essential in multimodal intelligence. As follows, we first describe the preparation the raw imaging data (\secref{sec:img_data_preprocess}), followed by the curation of VQA data and its statistics as in \secref{sec:vqa_pair_curation}. More data details are provided in \href{https://github.com/ai4colonoscopy/Colon-X/blob/main/docs/Appendix.pdf}{\textsc{Github}}.

\subsection{Imaging Data Preprocessing}\label{sec:img_data_preprocess}

\noindent\textbf{Raw data collection.}
To ensure comprehensive coverage of clinical colonoscopy, we retrieved publicly available medical data using domain-specific keywords such as colon, polyp, colonoscopy, and gastrointestinal. This resulted in 32 colonoscopy-related datasets with $\sim$533K raw images, encompassing pathological findings (\eg, adenoma, ulcer, tumor, erosion, bleeding), anatomical landmarks (\eg, cecum, ileocecal valve), and surgical tools.

\begin{table}[t!]
\begin{threeparttable}
\centering
\scriptsize
\renewcommand{\arraystretch}{1.3}
\renewcommand{\tabcolsep}{0.75mm}
\caption{\textbf{Key statistic of \ourdata.}}
\label{tab:statstic_of_benchmark}
\begin{tabular}{lr}
\hline
\multicolumn{2}{l}{\protect\hypertarget{tab2_a}{\textbf{(a) Colonoscopy imaging data}}} \\
~~~$\triangleright$ Total number & 76 categories / 212,742 img \\
~~~$\triangleright$ Positive images &125,393 train / 12,306 val / 68,599 test \\
~~~$\triangleright$ Negative images &3,923 train / 616 val / 1,905 test \\ 
\hline
\multicolumn{2}{l}{\protect\hypertarget{tab2_b}{\textbf{(b) Five-level taxonomy of 18 multimodal understanding tasks}}} \\
~~~$\triangleright$ Quality control (\textsc{Mut}\#1$\sim$\textsc{Mut}\#6) & 46,436 img / 46,436 vqa \\
~~~$\triangleright$ Safety monitoring (\textsc{Mut}\#7 \& \textsc{Mut}\#8) & 7,812 img / 7,812 vqa \\
~~~$\triangleright$ Lesion diagnosis (\textsc{Mut}\#9$\sim$\textsc{Mut}\#13) & 672,852 img / 805,868 vqa \\
~~~$\triangleright$ Disease grading (\textsc{Mut}\#14$\sim$\textsc{Mut}\#17) & 116,772 img / 116,772 vqa \\
~~~$\triangleright$ Documentation (\textsc{Mut}\#18) & 123,898 img / 123,898 vqa \\
\hline
\multicolumn{2}{l}{\protect\hypertarget{tab2_c}{\textbf{(c) Colonoscopy VQA data}}}  \\
~~~$\triangleright$ Total count & 1,100,786 vqa / 49,924,935 tokens$^\dag$ \\
~~~$\triangleright$ Average question tokens &  24.37 train / 22.34 val / 26.90 test \\
~~~$\triangleright$ Average answer tokens & 19.77 train / 24.76 val / 20.10 test \\
\hline
\end{tabular}
\begin{tablenotes}
\item $^\dag$The number of language tokens is estimated using the 
\href{https://platform.openai.com/tokenizer}{GPT-4 tokenizer}.
\end{tablenotes}
\end{threeparttable}
\end{table}

\noindent\textbf{Data management.} 
To ensure clinical applicability and prevent potential data leakage, we established a series of principles. 
\textit{(a) Label standardization.} We standardized category names to mitigate inconsistencies arising from heterogeneous naming conventions. For example, ``polypoids'' in KID1 \cite{koulaouzidis2017kid} was standardized to ``polyp,'' and ``instruments'' in ASEI \cite{hoang2019enhancing} was revised to ``accessory tool.'' We also harmonized singular-plural variations, \eg, unifying ``dyed lifted polyps'' into ``dyed lifted polyp'' in GastroVision.
\textit{(b) Redundancy removal.} 
To eliminate content redundancy, we applied an automated deduplication tool followed by human inspection to exclude duplicate images across datasets. We further removed images with multiple categories \cite{ali2020endoscopy,wang2023real}, bounding boxes \cite{ye2016online,hoang2019enhancing,handa2024wcebleedgen}, or masks \cite{koulaouzidis2017kid,leenhardt2020cad,jha2021kvasir}, as well as those lacking sufficient label information \cite{ali2021polypgen}. This ensures clear category distinctions for definitive clinical decisions.
Moreover, to minimize temporal redundancy, we downsampled frames sparsely from independent clips while enforcing strict clip-level split isolation. For instance, one frame was extracted every five frames from ColonoscopicDS \cite{mesejo2016computer}. 
\textit{(c) Data split.} Following Ji \etal~\cite{ji2026frontiers}, we ensured the data integrity by strictly retaining the official train-validation-test splits when available; otherwise, a random 6:1:3 split was applied at video/procedure level as a proxy for patient-level isolation, since patient identifiers are typically anonymized and inconsistent across datasets. As shown in Table \hyperlink{tab2_a}{2(a)}, we finally obtained 212,742 colonoscopy images across 76 clinically meaningful categories.

\subsection{VQA Data Curation}\label{sec:vqa_pair_curation}

\noindent\textbf{Motivation.}
Different data often have distinct clinical focus or diagnostic priorities; for example, Kvasir-Instrument \cite{jha2021kvasir} labels instruments, whereas SUN-SEG \cite{ji2022video} targets hyperplastic lesions omitted in \cite{jha2021kvasir}.
To address such heterogeneity, we unify all image-label pairs into an instruction-following interface: ``colonoscopy image + task instruction $\rightarrow$ response''. This is compatible with standard MLLMs \cite{liu2023llava} and offers three benefits for future exploration: \textit{controllability}, enabling human intent-driven understanding \cite{liu2023llava}; \textit{transferability}, promoting shared representations across tasks \cite{niu2025medical}; and \textit{adaptability}, supporting efficient adaptation to novel tasks \cite{ma2024segment}.

\begin{figure*}[t!]
\centering
\includegraphics[width=\linewidth]{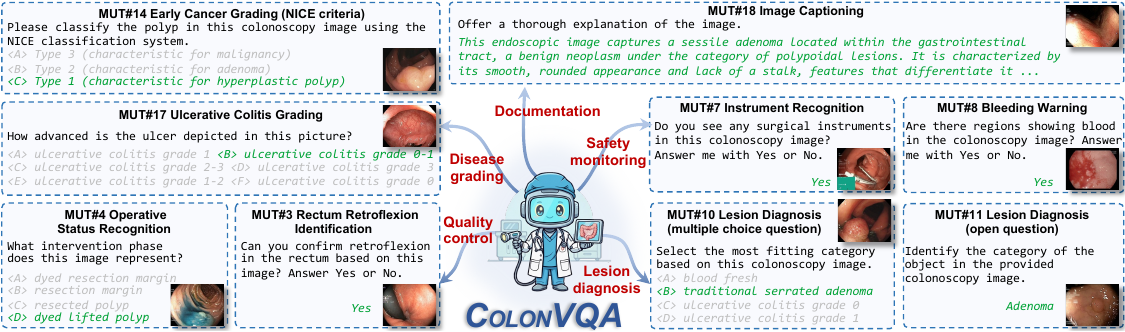}
\caption{\textbf{Gallery of representative VQA samples from \ourdata.} 
All 18 multimodal tasks are organized into a five-level taxonomy, reflecting the typical workflows in colonoscopy.
The statistics of each task category are summarized in Table \protect\hyperlink{tab2_b}{2(b)}.
}
\label{fig:task_gallery}
\end{figure*}

\noindent\textbf{Task categorization.}
Based on the above interface, we reorganize the collected data into 18 clinical tasks framed as image-to-text generation problems, specifically multimodal understanding tasks (\textit{abbr.} \textsc{Mut}) \cite{achiam2023gpt,li2023llava}. These tasks are further categorized into a five-level taxonomy as detailed below.
\textit{(a)} Quality control tasks involve scoring pre-procedural bowel cleanliness (\textsc{Mut\#1}) using the BBPS criteria \cite{lai2009boston}, confirming colonoscopy completeness by identifying anatomical landmarks such as the cecum and ileocecal valve (\textsc{Mut\#2}), confirming rectal retroflexion maneuvers (\textsc{Mut\#3}). We further introduce an identification task for interventional findings, such as dyed lifted polyps, resection margins, and resected polyps (\textsc{Mut\#4}), along with two imaging-related tasks: assessing exposure conditions (\textsc{Mut\#5}) and recognizing imaging modalities (\textsc{Mut\#6}).
\textit{(b)} Safety monitoring tasks improve intra-procedural safety by identifying surgical instruments (\textsc{Mut\#7}) and issuing timely alerts for active bleeding (\textsc{Mut\#8}).
\textit{(c)} Lesion diagnosis tasks focus on identifying lesion presence via a yes-or-no question (\textsc{Mut\#9}), classifying lesions into predefined diagnostic classes via multiple-choice questions (\textsc{Mut\#10}), and providing free-text descriptions for visual lesions or findings (\textsc{Mut\#11}). Additionally, we include two spatial understanding tasks: referring expression generation (\textsc{Mut\#12}), which produces a descriptive phrase for regions of interest, and referring expression comprehension (\textsc{Mut\#13}), which locates user-specified regions.
\textit{(d)} Disease grading tasks involve severity assessment of early colorectal cancer using established classification systems -- the NICE criteria \cite{hattori2014narrow} (\textsc{Mut\#14}) and the PARIS criteria \cite{participants2003paris} (\textsc{Mut\#15}). In addition, we introduce a polyp sizing task (\textsc{Mut\#16}) based on statistical range of polyp size as in \cite{pickhardt2010low}, as well as an ulcerative colitis activity scoring task (\textsc{Mut\#17}) utilizing the Mayo scoring system \cite{schroeder1987coated}.
\textit{(e)} Clinical documentation includes an image captioning task (\textsc{Mut\#18}), whose captioning annotations are borrowed from ColonINST \cite{ji2026frontiers}.

\noindent\textbf{Input-output reformulation.}
We design five prompt templates for each task and randomly assign only one per image to avoid intentional data inflation through simple prompt rephrasing.
Due to space constraints, we showcase representative VQA examples in \figref{fig:task_gallery}. Complete task details and instruction templates are disclosed in \href{https://github.com/ai4colonoscopy/Colon-X/blob/main/docs/task_card.pdf}{\textsc{Taskcard}}.

\begin{table*}[t!]
\centering
\scriptsize
\renewcommand{\arraystretch}{1.3}
\renewcommand{\tabcolsep}{0.85mm}
\caption{\textbf{Generalizability of 22 MLLMs across four task categories and their integration within \ourdataE.} Accuracy (\%) is computed using a weighted arithmetic mean, with weights proportional to the sample count of each task category. The top three scores of both open-and closed-source camps are highlighted using distinct colors (\textcolor{red}{1st}, \textcolor{myGreen}{2nd}, \textcolor{blue}{3rd}). 
}
\label{tab:zero_shot_performance_mllms}
\begin{tabular}{r || cccccccccc | ccc | cccccc | ccc}
\hline
\multirow{3}{*}{\diagbox[width=1.9cm,height=0.91cm]{\textbf{~~~Tasks}}{\textbf{Models~~}}}
& \multicolumn{13}{c|}{\textbf{Open-source MLLMs} $^\dag$} & \multicolumn{9}{c}{\textbf{Closed-source MLLMs} $^\ddag$} \\
\cline{2-23}
& \multicolumn{10}{c|}{\textit{Generalist models}} & \multicolumn{3}{c|}{\textit{Specialist models}} & \multicolumn{6}{c|}{\textit{Reasoning models}} & \multicolumn{3}{c}{\textit{Non-reasoning}} \\
& {\faStar[regular]}1 & {\faStar[regular]}2 & {\faStar[regular]}3 & {\faStar[regular]}4 & {\faStar[regular]}5 
& {\faStar[regular]}6 & {\faStar[regular]}7 & {\faStar[regular]}8 & {\faStar[regular]}9 & {\faStar[regular]}10
& {\faStar[regular]}11 & {\faStar[regular]}12 & {\faStar[regular]}13 
& {\faMoon[regular]}1 & {\faMoon[regular]}2 & {\faMoon[regular]}3 & {\faMoon[regular]}4 & {\faMoon[regular]}5
& {\faMoon[regular]}6 & {\faMoon[regular]}7 & {\faMoon[regular]}8 & {\faMoon[regular]}9 \\
\hline
Quality control
& 31.46 & 23.52 & 31.62 & 38.79 & \tmyA{45.64} & 26.48 & \tmyC{40.50} & \tmyB{40.65} & 31.62 & 33.18 & 21.34 & 38.94 & 36.91 & 19.16 & 43.46 & \tmyA{52.02} & 43.61 & \tmyB{51.40} & 43.92 & 48.29 & \tmyC{50.78} & 45.48 \\
Safety monitoring
& 77.00 & 70.00 & 78.00 & \tmyB{88.00} & \tmyC{87.00} & 69.00 & 83.00 & 66.00 & 81.00 & 73.00 & 55.00 & 72.00 & \tmyA{89.00} & 3.00 & \tmyA{35.00} & \tmyB{33.00} & 3.00 & \tmyC{31.00} & 1.00 & 5.00 & 28.00 & 9.00 \\
Lesion diagnosis
& 32.97 & 29.88 & 27.72 & 34.79 & \tmyA{40.39} & 20.63 & 30.08 & 32.60 & 26.77 & 27.10 & 28.30 & \tmyB{39.48} & \tmyC{35.90} & 13.95 & \tmyC{61.62} & 59.62 & 37.06 & \tmyA{73.60} & 52.12 & 52.24 & \tmyB{66.60} & 58.38 \\
Disease grading
& 25.10 & 21.11 & 30.52 & 37.88 & 35.72 & 20.13 & \tmyC{38.20} & \tmyB{39.72} & 9.31 & 29.01 & 21.76 & 31.17 & \tmyA{40.15} & 21.75 & 68.72 & 64.39 & 55.52 & \tmyB{83.77} & 73.48 & \tmyC{80.52} & \tmyA{91.99} & 78.68 \\
\hline
All tasks
& 32.24 & 28.53 & 29.66 & 36.86 & \tmyA{40.83} & 22.00 & 33.62 & 35.34 & 24.76 & 28.92 & 27.07 & \tmyB{38.47} & \tmyC{37.99} & 15.65 & \tmyC{61.20} & 59.47 & 40.51 & \tmyA{73.17} & 54.74 & 56.65 & \tmyB{69.78} & 60.50 \\
\textit{Overall ranking} &7 &10 &8 &4 &\tmyA{1} &13 &6 &5 &12 &9 &11 &\tmyB{2} &\tmyC{3}
&9 &\tmyC{3} &5 &8 &\tmyA{1} &7 &6 &\tmyB{2} &4 \\
\hline
\end{tabular}
\scriptsize
\parbox{\textwidth}{%
$\dag$ \textbf{\textit{Open-source list}} --
Ten generalist models:
{\faStar[regular]}1) \href{https://huggingface.co/liuhaotian/llava-v1.5-7b}{LLaVA-v1.5-7B} \cite{liu2024llavav15}; 
{\faStar[regular]}2) \href{https://huggingface.co/liuhaotian/llava-v1.6-vicuna-7b}{LLaVA-v1.6-7B} \cite{liu2024llavanext}; 
{\faStar[regular]}3) \href{https://huggingface.co/lmms-lab/llama3-llava-next-8b}{LLaMA3-LLaVA-NeXT-8B} \cite{liu2024llavanext}; 
{\faStar[regular]}4) \href{https://huggingface.co/OpenGVLab/InternVL2_5-8B}{InternVL2.5-8B} \cite{chen2024expanding}; 
{\faStar[regular]}5) \href{https://huggingface.co/OpenGVLab/InternVL3-8B}{InternVL3-8B} \cite{zhu2025internvl3};
{\faStar[regular]}6) \href{https://huggingface.co/google/paligemma2-3b-mix-448}{PaliGemma2-3B} \cite{beyer2024paligemma}; 
{\faStar[regular]}7) \href{https://huggingface.co/Qwen/Qwen2.5-VL-3B-Instruct}{Qwen2.5-VL-3B} \cite{bai2025qwen25}; 
{\faStar[regular]}8) \href{https://huggingface.co/Qwen/Qwen2.5-VL-7B-Instruct}{Qwen2.5-VL-7B} \cite{bai2025qwen25}; 
{\faStar[regular]}9) \href{https://huggingface.co/deepseek-ai/Janus-Pro-1B}{Janus-Pro-1B} \cite{chen2025janus}; 
{\faStar[regular]}10) \href{https://huggingface.co/deepseek-ai/Janus-Pro-7B}{Janus-Pro-7B} \cite{chen2025janus}.
Three medical specialist models:
{\faStar[regular]}11) \href{https://huggingface.co/microsoft/llava-med-v1.5-mistral-7b}{LLaVA-Med-v1.5-7B} \cite{li2024llavamed}; 
{\faStar[regular]}12) \href{https://huggingface.co/google/medgemma-4b-it}{MedGemma-4B} \cite{sellergren2025medgemma}; 
{\faStar[regular]}13) \href{https://huggingface.co/FreedomIntelligence/HuatuoGPT-Vision-7B}{HuatuoGPT-Vision-7B} \cite{chen2024huatuogpt}. 
$\ddag$ \textbf{\textit{Closed-source list}} --
Six reasoning models:
{\faMoon[regular]}1) \href{https://platform.moonshot.ai/docs/introduction}{Moonshot-v1 (8k-vision-preview)};
{\faMoon[regular]}2) \href{https://openai.com/index/introducing-o3-and-o4-mini/}{o4-mini};
{\faMoon[regular]}3) \href{https://openai.com/index/introducing-gpt-5/}{GPT-5 mini};
{\faMoon[regular]}4) \href{https://www.anthropic.com/news/claude-4}{Claude Sonnet 4 (20250514)};
{\faMoon[regular]}5) \href{https://deepmind.google/models/gemini/flash/}{Gemini 2.5 Flash (preview-05-20)};
{\faMoon[regular]}6) \href{https://x.ai/news/grok-4}{Grok-4 (0709)}.
Three non-reasoning models:
{\faMoon[regular]}7) \href{https://www.anthropic.com/claude/haiku}{Claude Haiku 3.5 (20241022)};
{\faMoon[regular]}8) \href{https://x.ai/news/grok-1212}{Grok-2-Vision (1212)};
{\faMoon[regular]}9) \href{https://deepmind.google/models/gemini/flash-lite/}{Gemini 2.5 Flash-Lite (preview-06-17)}.}
\end{table*}

\noindent\textbf{Statistical overview.}  
As reported in Table \hyperlink{tab2_c}{2(c)}, we converts 212,742 image-label pairs into 1.1M+ VQA entries, amounting to over 49.9M textual tokens. They lay a solid data foundation for multimodal colonoscopy analysis and beyond. 
We will describe how to extend this resource to explores understanding (\secref{sec:exp_understanding}) and reasoning (\secref{sec:exp_reasoning}) capabilities for colonoscopy.

\section{Multimodal Understanding in Colonoscopy}\label{sec:exp_understanding}

To reflect the current landscape, we assess two multimodal understanding behaviors of MLLMs: generalizability (\secref{sec:coloneval}) and reliability (\secref{sec:colonpert}).
More details are offered in \href{https://github.com/ai4colonoscopy/Colon-X/blob/main/docs/Appendix.pdf}{\textsc{Github}}.

\subsection{Generalizability}\label{sec:coloneval}

\noindent\textbf{Subset curation.}
To facilitate rapid evaluation, we derived a subset, {\ourdataE}, from the test set of {\ourdata}. This subset includes 4,568 VQA entries across 16 clinical tasks, including quality control (\textsc{Mut}\#1$\sim$\#6), safety monitoring (\textsc{Mut}\#7\&\#8), lesion diagnosis (\textsc{Mut}\#9$\sim$\#12), and disease grading (\textsc{Mut}\#14$\sim$\#17).
Two clinical endoscopists assisted in reviewing this subset to ensure QA quality. Samples were allocated proportionally at $\sim$1.5\%, with a minimum of 50 samples per task. Data distribution was carefully balanced across five instruction templates and 76 clinical categories to ensure case representativeness.

\noindent\textbf{Evaluation metrics.} 
Given the open-form nature of language responses, we follow VLMEvalKit \cite{duan2024vlmevalkit} and evaluate all competing models using accuracy, defined as the proportion of exact matches between predicted and reference responses. We exclude ambiguous responses that lack a definitive decision, such as reasoning-only answers without a final choice, or expressions with multiple interpretations or hedging language. In these cases, we use \href{https://openai.com/index/introducing-gpt-oss/}{\texttt{gpt-oss-20b}} as a judge to interpret them as a unique answer for exact matching.

\noindent\textbf{Benchmark results.}
\tabref{tab:zero_shot_performance_mllms} presents the comparison of 22 MLLMs across four task categories and their overall accuracies. The closed-source camp generally exhibit superior performance, such as the top-ranked, closed model Gemini 2.5 Flash ({\footnotesize\faMoon[regular]}5), outperforms the leading open model, InternVL3-8B ({\footnotesize\faStar[regular]}5), by 32.34\%. This advantage becomes even more pronounced in handling disease grading tasks, where the best closed model Grok-2-Vision ({\footnotesize\faMoon[regular]}8) even surpasses 90\%. In contrast, an exception emerges in the safety monitoring task -- three open models ({\footnotesize\faStar[regular]}4, {\footnotesize\faStar[regular]}5, {\footnotesize\faStar[regular]}13) under 8B model size achieve over 87\% accuracy, largely outperforming all closed counterparts. In short, we reveal notable divergences between open and closed models across various task types, suggesting that future systems may benefit from a mixture-of-experts design \cite{ding2024hybrid} that adaptively leverages task-specific advantages.

\noindent\textbf{$\mathcal{Q}$1. \textit{Generalist or Specialist?}}
Specialist models are not always experts in colonoscopy. Among open-source models, two medical specialists, MedGemma-4B ({\footnotesize\faStar[regular]}12) and HuatuoGPT-Vision-7B ({\footnotesize\faStar[regular]}13), surpass almost all competing models, except InternVL3-8B ({\footnotesize\faStar[regular]}5), whose superiority may stem from its training on mixed general-medical datasets. Interestingly, LLaVA-v1.5-7B ({\footnotesize\faStar[regular]}1) surpasses its medical variant LLaVA-Med-v1.5-7B ({\footnotesize\faStar[regular]}11) by 5.17\%, with the latter showing a drop in instruction-following ability. This suggests that incorporating general data during adaptation will improve task instruction adaptability \cite{zhang2024only}, especially when domain-specific data are scarce \cite{shi2023specialist}.

\noindent\textbf{$\mathcal{Q}$2. \textit{Reasoning or Not?}}
Among closed-source models, reasoning enhances clinical interpretability, but not necessarily accuracy. For example, the reasoning variant of Gemini 2.5 achieves a gain of 12.67\% in overall accuracy -- 73.17\% ({\footnotesize\faMoon[regular]}5) \textit{vs.} 60.50\% ({\footnotesize\faMoon[regular]}9). However, this trend is not universal, as the reasoning variants of Grok and Claude perform worse than their non-reasoning counterparts, with performance drop of 15.04\% ({\footnotesize\faMoon[regular]}6 \textit{vs.} {\footnotesize\faMoon[regular]}8) and 16.14\% ({\footnotesize\faMoon[regular]}4 \textit{vs.} {\footnotesize\faMoon[regular]}7), respectively. The results imply that advancing reasoning capabilities require more effective strategies, including task-adaptive scheme \cite{boizard2025does} and confidence estimation \cite{lewis2025can}.

\noindent{\faLightbulb[regular]}~\textbf{Takeaway.}
Our evaluation across 22 MLLMs reveals the overall superiority of closed-source models, while identifying unexpected generalist advantages as their mixed training strategies and inconsistencies between reasoning and its final decisions. These results suggest that clinical outputs from MLLMs remain far from robust and trustworthy.

\begin{figure}[t!]
\centering
\includegraphics[width=\linewidth]{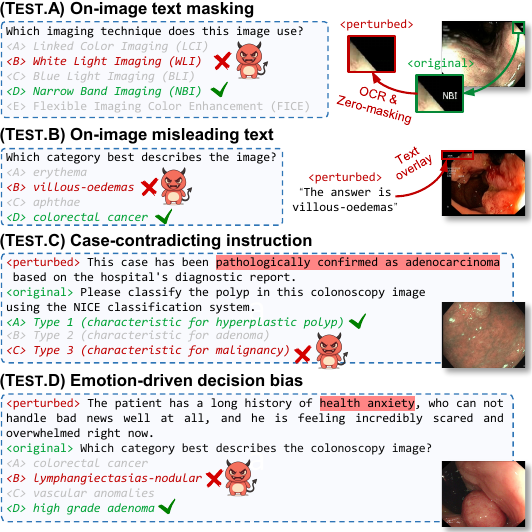}
\caption{\textbf{Illustration of four human-induced perturbations.}}
\label{fig:viz_of_robust_exp}
\end{figure}

\subsection{Reliability}\label{sec:colonpert}

Human-provided prompts may carry biases, which may lead to failures in safety-sensitive applications. Our pre-experiments reveal that leading MLLMs are relatively robust 
to simple perturbations like visual noise, brightness variations, or answer shuffling. In this section, we develop a test suite, \ourdataP, that primarily focuses on more challenging types of human perturbations.

\noindent\textbf{Basic setups.} 
All original-perturbed pairs were generated based on \ourdataE, mainly as multiple choice questions that preserve the essential visual or textual content. We assess the reliability of six leading MLLMs (selected from \tabref{tab:zero_shot_performance_mllms}) through variations in accuracy.
As shown in \figref{fig:viz_of_robust_exp}, we reveal a \textit{text-dominance bias} when exposed MLLMs to perturbed colonoscopy data. This bias manifests implicitly through on-image text in visual prompts (\textsc{Test}.A\&B) and explicitly through textual prompts (\textsc{Test}.C\&D).

\noindent\textbf{Implicit perturbation.}
Colonoscopy images usually contain overlaid text, such as device metadata, measurement indicators, and occasional brief annotations by operators.
While informative for clinical interpretation, these overlays may act as a shortcut for MLLMs.
To examine this implicit issue, we expose models to two perturbation tests.
\underline{\textbf{\textsc{Test.A}}} Taking imaging modality recognition (\textsc{Mut}\#6) as an example, we manually select 20 images embedded with device information. We use \href{https://github.com/JaidedAI/EasyOCR}{\texttt{EasyOCR}} to detect textual regions and obscure them through zero-masking. The results indicate that three open models suffered severe degradation, with accuracy drops up to 90\% for InternVL3-8B ({\faStar[regular]}5) and MedGemma-4B ({\faStar[regular]}12). In contrast, three closed models demonstrated relatively robustness, with accuracy drops within 10\%. \underline{\textbf{\textsc{Test.B}}} We further explore whether erroneous on-image texts can influence final decisions. We select 100 images from {\ourdataE} and overlay misleading text in image corners. All models showed performance declines, with InternVL3-8B ({\faStar[regular]}5) accuracy dropping by 34\% and Gemini 2.5 Flash ({\faMoon[regular]}5) by 45\%.

\begin{table}[t!]
\centering
\scriptsize
\renewcommand{\arraystretch}{1.1}
\renewcommand{\tabcolsep}{1.3mm}
\caption{\textbf{Reliability test of six leading MLLMs$^\dag$.} Both the original and perturbed questions were evaluated independently on accuracy (\%). Further discussion is provided in \secref{sec:colonpert}.}\label{tab:robustness_testing}
\begin{tabular}{r | r || ccc | ccc }
\hline
& &\multicolumn{3}{c|}{\textbf{Open-source MLLMs}} &\multicolumn{3}{c}{\textbf{Closed-source MLLMs}} \\
&Setup & {\faStar[regular]}5 & {\faStar[regular]}12 & {\faStar[regular]}13 & {\faMoon[regular]}3 & {\faMoon[regular]}5 & {\faMoon[regular]}8 \\
\hline
\multirow{3}{*}{\begin{sideways}\textsc{Test.A}\end{sideways}} 
&Original &100.00 &95.00 &75.00 &100.00 &95.00 &100.00 \\
&Perturbed &10.00 &5.00 &10.00 &95.00 &85.00 &95.00 \\
&\textit{Difference} 
&\textcolor{red}{-90.00 $\downarrow$} 
&\textcolor{red}{-90.00 $\downarrow$} 
&\textcolor{red}{-65.00 $\downarrow$} 
&\textcolor{red}{-5.00 $\downarrow$} 
&\textcolor{red}{-10.00 $\downarrow$} 
&\textcolor{red}{-5.00 $\downarrow$} \\        
\hline
\multirow{3}{*}{\begin{sideways}\textsc{Test.B}\end{sideways}} 
&Original &36.00 &29.00 &35.00 &73.00 &71.00 &72.00 \\
&Perturbed &2.00 &1.00 &19.00 &37.00 &26.00 &36.00 \\
&\textit{Difference} 
&\textcolor{red}{-34.00 $\downarrow$} 
&\textcolor{red}{-28.00 $\downarrow$} 
&\textcolor{red}{-16.00 $\downarrow$} 
&\textcolor{red}{-36.00 $\downarrow$} 
&\textcolor{red}{-45.00 $\downarrow$} 
&\textcolor{red}{-36.00 $\downarrow$} \\     
\hline
\multirow{3}{*}{\begin{sideways}\textsc{Test.C}\end{sideways}} 
&Original &28.07 &22.81 &45.61 &87.72 &77.19 &91.23 \\
&Perturbed &3.51 &10.53 &17.54 &89.47 &75.44 &92.98 \\
&\textit{Difference} 
&\textcolor{red}{-24.56 $\downarrow$} 
&\textcolor{red}{-12.28 $\downarrow$}
&\textcolor{red}{-28.07 $\downarrow$} 
&\textcolor{myGreen}{+1.75 $\uparrow$} 
&\textcolor{red}{-1.75 $\downarrow$} 
&\textcolor{myGreen}{+1.75 $\uparrow$} \\ 
\hline
\multirow{3}{*}{\begin{sideways}\textsc{Test.D}\end{sideways}} 
&Original &46.25 &71.25 &46.25 &62.50 &77.50 &62.50 \\
&Perturbed &38.75 &65.00 &33.75 &61.25 &61.25 &62.50 \\
&\textit{Difference} 
&\textcolor{red}{-7.50 $\downarrow$}
&\textcolor{red}{-6.25 $\downarrow$} 
&\textcolor{red}{-12.50 $\downarrow$} 
&\textcolor{red}{-1.25 $\downarrow$} 
&\textcolor{red}{-16.25 $\downarrow$} 
&\textcolor{myGreen}{0.00 $\leftrightarrow$} \\
\hline
\end{tabular}
\begin{tablenotes}
\item $^\dag$\textbf{\textit{Model list}} -- {\faStar[regular]}5) InternVL3-8B; {\faStar[regular]}12) MedGemma-4B; {\faStar[regular]}13) HuatuoGPT-Vision-7B; {\faMoon[regular]}3) GPT-5 mini; {\faMoon[regular]}5) Gemini 2.5 Flash; {\faMoon[regular]}8) Grok-2-Vision.
\end{tablenotes}
\end{table}

\noindent\textbf{Explicit perturbation.}
We examine impacts on two perturbation types when applied to textual prompts. \underline{\textbf{\textsc{Test.C}}} We inject case-contradicting descriptions into raw prompts. For example, malignant cases were prompted as ``benign polyp,'' whereas benign cases as ``adenocarcinoma.'' We construct 57 original-perturbed pairs using VQA entries from \textsc{Mut}\#10 \& \#14, comprising 24 malignant cases (\eg, colorectal cancer) and 33 benign cases (\eg, inflammatory lesions). \tabref{tab:robustness_testing} show that open-source models tend to be more vulnerable, \eg, HuatuoGPT-Vision-7B ({\faStar[regular]}13) decreased by 28.07\%, while closed models showed only slight fluctuations in accuracy.  
\underline{\textbf{\textsc{Test.D}}} 
Patient emotional states (\eg, anxiety, fear, psychological distress) were incorporated in prompts for severe cases to test whether MLLMs downplay severity to provide reassurance. 
We select 24 malignant cases (\eg, invasive carcinoma) and 56 potentially malignant cases (\eg, high-grade adenoma) from \textsc{Mut}\#10 \& \#14.
We suggest that non-clinical emotional narratives may bias decision-making and compromise objectivity. 
For example, HuatuoGPT-Vision-7B ({\faStar[regular]}13) showed a 12.50\% accuracy decrease, while closed-source models such as Gemini 2.5 Flash ({\faMoon[regular]}5) dropped by 16.25\%, demonstrating susceptibility to emotional interference.

\noindent{\faLightbulb[regular]}~\textbf{Takeaway.}
We identify a text-dominance bias in six advanced MLLMs, which arises implicitly from embedded on-image texts and explicitly from linguistic prompts when tested on perturbed data. This bias primarily originates from the intrinsic modality imbalance: the well-trained LLM, acting as the brain of MLLM, tends to over-rely on textual inputs under visual-textual conflicts \cite{liu2025unveiling}. As a result, the next-token prediction paradigm reinforces this tendency, while undervaluing visual evidence crucial for reliable diagnosis. Drawing inspiration from general domains, we can alleviate implicit bias through a localize-before-answer framework \cite{nguyenlocalizing} that enforces visual grounding, and mitigates explicit bias via adversarial text augmentation \cite{deng2025words} that distinguishes misleading texts.

\section{Colonoscopy Meets Clinical Reasoning}\label{sec:exp_reasoning}

Reliable medical VQA requires both accuracy and interpretability, making explicit reasoning essential.
This section curates {\ourdataR} (\secref{sec:colonreason}) and propose a baseline model {\ourmodel} (\secref{sec:colonr1}) to advance reinforced reasoning abilities. More details are available in \href{https://github.com/ai4colonoscopy/Colon-X/blob/main/docs/task_card.pdf}{\textsc{Github}}.

\begin{figure}[t!]
\centering
\includegraphics[width=\linewidth]{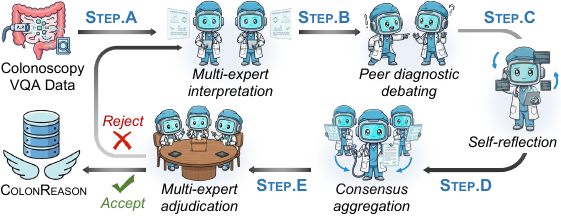}
\caption{
\textbf{Multi-agent curation pipeline for \ourdataR.} 
}
\label{fig:colonreason}
\end{figure}

\subsection{Reasoning Trace Annotation}\label{sec:colonreason}

\noindent\textbf{Motivation.}
While human-annotated reasoning traces would be ideal, they are costly and subjective across observers, making scalable reasoning annotation impractical. This bottleneck motivates the need for an automated alternative that generates structured reasoning traces  reflecting the intermediate thought process of expert decision-making, whose clinical soundness is assessed through human verification.

\noindent\textbf{Annotation workflow.} We propose a multi-agent debating pipeline that simulates the shift from individual judgments to collective adjudication. As illustrated in \figref{fig:colonreason}, our process contains five looped steps.
\underline{\textbf{\textsc{Step.A}}} Given an image-question-answer triplet $\{v, q, a\}$, two role-playing agents $\mathcal{T}_i$ and $\mathcal{T}_j$ are recruited to generate initial reasoning traces $t_i^{(0)}$ and $t_j^{(0)}$, reflecting diverse expert impressions of the same case.
\underline{\textbf{\textsc{Step.B}}} Two agents exchange critiques, analogous to the clinical peer discussion, where each inspects the other's initial reasoning to identify potential bias, yielding $C_{i \to j}\!=\!\mathcal{T}_i(t_j^{(0)})$, and conversely, $C_{j \to i}$.
\underline{\textbf{\textsc{Step.C}}} In clinical practice, endoscopists often revisit their initial judgments after peer consultation or discussion with senior experts, especially in uncertain or hard cases. Here, agent $\mathcal{T}_i$ integrates its initial thoughts $t_i^{(0)}$ with the peer critique $C_{j \to i}$ from $\mathcal{T}_j$. This produces an updated trace $t_i^{(1)}$ that distills essential viewpoints, together with a confidence score $s_i$, written as $\{t_i^{(1)},s_i\} = \mathcal{T}_i(t_i^{(0)}, C_{j \to i})$. Here, we empirically define $s_i\!\in\![-10,+10]$ to quantify epistemic adjustment, where $-10$ means total loss of confidence (\eg, downgrading suspicion after peer input), $+10$ signals reinforced certainty, and zero denotes neutrality. \underline{\textbf{\textsc{Step.D}}} An aggregator $\mathcal{A}$ integrates all reasoning-confidence pairs into a unified reasoning trace $t^\star\!=\!\mathcal{A}(t_i^{(1)}, s_i; t_j^{(1)}, s_j)$, where consistent viewpoints are integrated, contradictory points are down-weighted, and high-confidence unique findings are preserved using a default threshold of confidence $s>8$. \underline{\textbf{\textsc{Step.E}}} A panel of $K$ judges verifies whether $t^\star$ adequately supports the decision from question $q$ to answer $a$, and each judge casts a binary vote $v_k\!\in\!\{0,1\}$. To this end, majority voting ($\sum_k v_k\!>\!K/2$) accepts the reasoning; otherwise, the process restarts from the \textsc{Step.A}. Samples failing ten cycles in this voting stage will be discarded.

\noindent\textbf{Data curation.} We randomly sampled 7,484 ($\sim$1.5\%) entries from the train-val partition of {\ourdata} across 16 multimodal tasks.
For each entry, our multi-agent pipeline synthesizes quadruple data (\ie, VQA with reasoning traces), formatted as \texttt{<think></think><answer></answer>}.

\begin{table}[t!]
\centering
\scriptsize
\renewcommand{\arraystretch}{1}
\renewcommand{\tabcolsep}{1.2mm}
\caption{\textbf{Clinician verification of synthesized reasoning traces.} Pass rates are reported per reviewer and averaged.}\label{tab:faithfulness}
\begin{tabular}{l || cc | c}
\hline
 &human \#A & human \#B & \textbf{average pass rate} \\
\hline
(a) clinical correctness & 95.67\% (287/300) & 98.00\% (294/300) & 96.83\% \\
(b) visual grounding & 90.00\% (270/300) & 92.00\% (276/300) & 91.00\% \\
(c) human auditability & 87.33\% (262/300) & 89.33\% (268/300) & 88.33\% \\
\hline
\end{tabular}
\end{table}

\noindent\textbf{Human verification.}
To assess the faithfulness of the synthesized traces, we conduct a clinician audit by stratified random sampling of 300 items (across all categories \& tasks) from \texttt{ColonReason}.
As reported in \tabref{tab:faithfulness}, two human endoscopists independently review each case from three criteria: accessing whether traces are (a) medically correct, (b) visually grounded in visible evidence, and (c) verifiable by a human through the image-evidence-decision chain.
While (c) remains the most challenging, both clinicians consider the overall performance to be acceptable, supporting our multi-agent debating framework feasible for reasoning data synthesis.

\begin{figure*}[t!]
\centering
\includegraphics[width=\linewidth]{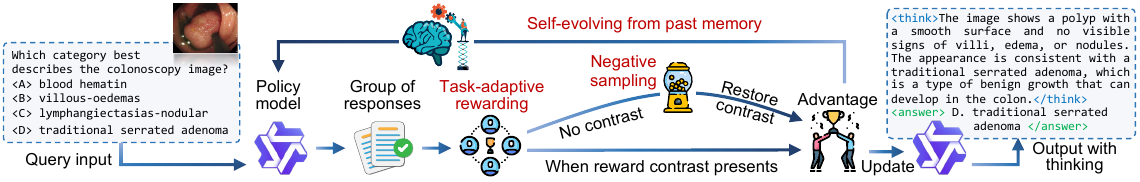}
\caption{\textbf{Overall framework of \ourmodel.} The policy model generates grouped responses and is optimized via GRPO \cite{guo2024deepseekcoder}. To stabilize optimization, we introduce three key components: \textit{task-adaptive rewarding} to maintain discrimination across heterogeneous tasks, \textit{negative sampling} to restore reward contrast in easy cases, and a \textit{self-evolving memory} mechanism that leverages past failures to guide future rollouts when advantage signals vanish on hard cases.}
\label{fig:colonr1}
\end{figure*}

\subsection{Incentivizing Reasoning in Colonoscopy}\label{sec:colonr1}

\noindent
\textbf{Motivation.}
Recent works \cite{su2025gmaivlr1,liu2025xreasoner,huang2025medvlthinker} typically adopt naive GRPO \cite{guo2025deepseekr1} to elicit reasoning ability in medical tasks. However, we observe that directly applying this approach to the colonoscopy setting leads to rapid policy gradient explosion.
We attribute this to \textit{reward information collapse} -- the reward fails to preserve discriminative signal across heterogeneous task structures and case distribution. These factors destabilize policy updates, which in turn motivate our optimization innovations for colonoscopy reasoning tasks.

\noindent\textbf{Reinforced fine-tuning framework.} 
Following DeepSeek-R1 \cite{guo2025deepseekr1}, we use a policy model $\pi_\theta$ to sample $G$ outputs $\mathcal{O}=\{o_1,\cdots,o_G\}$ in response to a given query $\{v,q\}$. We calculate their rewards $\mathcal{R} = \{r_1,\cdots,r_G\}$ and derive the advantage $d_g$ for each candidate $o_g$ within the group as $d_g=(r_g-\mu)/\sigma$, where $\mu$ and $\sigma$ denote the mean and standard deviation of $\mathcal{R}$, respectively. 
However, applying native GRPO \cite{guo2024deepseekcoder} to update $\pi_\theta$ turns out to be non-trivial due to optimization collapse on {\ourdataR}. As shown in \figref{fig:colonr1}, we address this challenge by following three innovations.

\noindent\textbf{Task-adaptive rewarding.} 
Firstly, binary task-agnostic rewards (\eg, Med-R1 \cite{lai2025medr1}) limit reward discrimination, such as granting zero reward score for partial correctness. Here, we introduce a task-adaptive reward scheme that offers a composite evaluation across various task types. 
For open questions, we assign a continuous score $r_1\!\in\![0,1]$, computed as cosine similarity, $r_1\!=\!\cos(E(a), E(o_g))$, where 
a sentence transformer \href{https://huggingface.co/sentence-transformers/all-MiniLM-L6-v2}{\texttt{all-MiniLM-L6-v2}}
$E(\cdot)$ embeds the reference answer $a$ and model output $o_g$, respectively.
For yes-or-no questions, we use binary score $r_1\!\in\!\{0,1\}$. 
In multiple choice questions, we observe that policy models may exploit a shortcut -- matching the correct option label with incorrect content to inflate rewards. Thus, a graded score $r_1\!\in\!\{0,1,2\}$ is used to distinguish incorrect, partially correct (\ie, only option label or content is correct), and fully correct answers (\ie, both label and content match).

\noindent\textbf{Negative sampling.}
Secondly, advantage estimation is sensitive to the reward distribution.
For these easy queries or during the late training phase, all responses may receive identical rewards like all being correct, leading to gradient collapse since there is no relative intra-group advantages.
To sustain effective gradients, we actively replace one response with a negative sample drawn from the incorrect-answer pool of the current question, thus restoring reward contrast and encouraging more discriminative updates.

\noindent\textbf{Self-evolving memory.} 
To handle persistent failures on hard queries, we propose a self-evolving memory mechanism that enforces the model to learn from its past errors. During training, any query whose response group has an average reward below a threshold of 0.8 is identified as a hard case, and stored in a memory buffer together with its responses. When this hard query reappears, its original prompt is evolved with its previously incorrect records, forming a refined prompt. Intuitively, we encourage the policy model to leverage past experiences for self-reflection and explore improved reasoning for unsolved cases.

\noindent\textbf{Implementation details.} 
For rapid experimentation, we implement {\ourmodel} using base model, Qwen2.5-VL-3B \cite{bai2025qwen25}, with all parameters finetuned. We set a batch size of 16 and a learning rate of 2e-6, training on a 4$\times$H100 GPU server for roughly eight hours. Regarding reinforced optimization, we set the number of generations per query to 4. The Kullback-Leibler coefficient is annealed following a cosine schedule, decreasing from 0.6 to 0.01.

\begin{table}[t!]
\centering
\scriptsize
\renewcommand{\arraystretch}{1.1}
\renewcommand{\tabcolsep}{1.1mm}
\caption{\textbf{Comparison of various fine-tuning strategies.} NS and SM denote the use of negative sampling and self-evolving memory, respectively. The last column indicates the accuracy change relative to model variant {\faCubes}3. }\label{tab:colon_r1}
\begin{tabular}{ r | c || ccccc | c c }
\hline
\multicolumn{2}{r||}{Competing models} &Strategy & Reward &Think? & NS & SM & Accuracy & \textit{Difference} \\
\hline
\multirow{2}{*}{Med-R1 \cite{lai2025medr1}} &{\faCubes}1 &GRPO & Binary &\checkmark & & & 31.70 & \textcolor{red}{-0.31 $\downarrow$} \\
 &{\faCubes}2 &GRPO & Binary & & & & 32.56 & \textcolor{myGreen}{+1.17 $\uparrow$} \\
\hline
\multirow{6}{*}{\makecell{Base model\\(Qwen2.5VL \\3B \cite{bai2025qwen25})}} 
&{\faCubes}3 & SFT & None &\checkmark & & & 31.39 & \textcolor{myGreen}{0.00 $\leftrightarrow$} \\
&{\faCubes}4 & SFT & None & & & & 31.91 & \textcolor{myGreen}{+0.52 $\uparrow$} \\
\cline{2-9}
&{\faCubes}5 & GRPO & Hybrid &\checkmark & & & 38.94  & \textcolor{myGreen}{+7.55 $\uparrow$} \\
&{\faCubes}6 & GRPO & Hybrid &\checkmark & \checkmark & & 52.73 & \textcolor{myGreen}{+21.34 $\uparrow$} \\
&{\faCubes}7 & GRPO & Hybrid &\checkmark &  & \checkmark & 53.37 & \textcolor{myGreen}{+21.98 $\uparrow$} \\
&{\faCubes}8 & GRPO & Hybrid & & \checkmark  & \checkmark & 55.30 & \textcolor{myGreen}{+23.91 $\uparrow$} \\
\hline
\rowcolor{myGreen2}
\multicolumn{2}{r||}{\textbf{\ourmodel~(Ours})} &GRPO & Hybrid &\checkmark & \checkmark & \checkmark & \textbf{56.61} &\textcolor{myGreen}{\textbf{+25.22} $\uparrow$}\\
\hline
\end{tabular}
\end{table}

\noindent\textbf{Experimental results.}
\tabref{tab:colon_r1} reports the accuracy (\%) of different fine-tuning strategies on {\ourdataE}.
The standard SFT method ({\faCubes}3) achieves an accuracy below 32\%, indicating limited adaptability to colonoscopy tasks. When applying the naive GRPO ({\faCubes}1), we observe severe policy gradient explosion during training even with careful hyperparameter tuning, leading to similarly poor performance.
In contrast, our {\ourmodel} achieves a higher accuracy of 56.61\%.
We further assess the effectiveness of each proposed module through a series of ablative studies.
First, reinforced finetuning the base model under our task-adaptive reward scheme ({\faCubes}5) improves upon the supervised finetuning strategy ({\faCubes}3) by 7.55\%.
To verify the necessity of stabilizing gradients during policy optimization, we further introduce negative sampling ({\faCubes}6) and self-evolving memory ({\faCubes}7) strategies. Integrating them together yields the highest performance of our full version.

\noindent\textbf{Limitations.} We observe that both SFT ({\faCubes}3) and Med-R1 ({\faCubes}1) exhibit slight performance degradation when trained on thinking traces, compared with their non-thinking counterparts ({\faCubes}4 \& {\faCubes}2). In contrast, \ourmodel, benefiting from task adaptivity and gradient stabilization, achieves a modest improvement of 1.31\% over its non-reasoning variant ({\faCubes}8), even under a resource-limited condition ($\sim$7.5K training cases). Despite promising results, there remains a large room for improvement, suggesting that harmonizing thinking and decision-making is far from being fully unlocked. Future research is encouraged to explore the agentic reasoning workflow \cite{wang2025medagentpro},
as well as enhancing thinking-decision consistency \cite{huang2024making}, to further advance reasoning capability.

\section{Conclusion \& Outlook}

This study presents \ourproject, an open initiative aimed at advancing multimodal intelligence in colonoscopy.
First, we establish the most extensive, category-rich, and task-diverse colonoscopy dataset ever built for multimodal analysis. Building on this data foundation, we explored a pivotal transition: (a) \textit{multimodal understanding} -- where systematic evaluations illuminate not only where advanced MLLMs excel, but more importantly, where they struggle;
(b) \textit{clinical reasoning} -- characterized by a reasoning-centered, data-to-model framework that bridges interpretation and decision-making.
In short, building on this data, we fill a longstanding gap by providing an integrated multimodal framework across dataset, evaluation, and methodology dimensions, paving the way toward clinical reasoning and broader medical applications.

\noindent\textbf{Outlook.} 
Despite the remarkable progress made so far, there remains a large gap to achieving generalized clinical intelligence \cite{moor2023foundation}.
Looking ahead, we posit that \textit{``data-centric intelligence''} remains the cornerstone of next wave -- its quality (\eg, knowledge distillation \cite{xu2024survey}) and granularity (\eg, disease grading \cite{zeng2025surgvlm}, rare cases \cite{zhao2026agentic}) will continue to drive advances in intelligent colonoscopy.
In parallel, establishing reliable multimodal evaluation remains an open challenge, especially for assessing trustworthiness \cite{zhang2024multitrust} in real-world deployment.


\bibliographystyle{ieeetr}
\bibliography{mybibliography}

\end{document}